\documentclass{article}

\usepackage{arxiv}

\usepackage[utf8]{inputenc} 
\usepackage[T1]{fontenc}    
\usepackage{hyperref}       
\usepackage{url}            
\usepackage{booktabs}       
\usepackage{amsfonts}       
\usepackage{nicefrac}       
\usepackage{microtype}      
\usepackage{lipsum}		
\usepackage{graphicx}
\usepackage{natbib}
\usepackage{doi}
\usepackage{graphicx}
\usepackage{xcolor}
\usepackage{algorithm}
\usepackage{algpseudocode}
\usepackage{tikz,amsmath} 
\usepackage{subcaption}

\title{Horizontal Federated Computer Vision}

\author{ 
\href{https://orcid.org/0000-0002-4966-0494}{\includegraphics[scale=0.06]{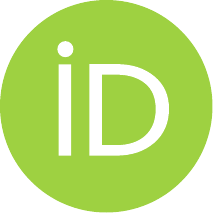}\hspace{1mm}Paul K. Mandal $^{1,*}$} \\
	\texttt{mandal@utexas.edu} \\
        \And
        \href{https://orcid.org/0009-0008-6875-6281}{\includegraphics[scale=0.06]{orcid.pdf}\hspace{1mm}Cole Leo $^{2}$} \\
        \texttt{cleo@t2s-solutions.com}
	\And
    Connor Hurley $^{2}$\\
	\texttt{churley@t2s-solutions.com} \\
    \And
    \\
    $^1$ The University of Texas at Austin, Austin TX, USA \\
    $^2$ T2S-Solutions, Aberdeen MD, USA \\
    $^*$ Corresponding Author \\
}

\renewcommand{\headeright}{}
\renewcommand{\undertitle}{}

\hypersetup{
pdftitle={Horizontal Federated Computer Vision},
pdfsubject={cs.LG, cs.CV, cs.NE},
pdfauthor={Paul K. Mandal, Cole Leo, and Connor Hurley},
pdfkeywords={computer vision, convolutional network (CNN), deep learning, federated learning, FCN (Fully Convolutional Network), FRCNN (Federated Faster R-CNN), image segmentation, machine learning, object detection, object recognition},
}

\begin{document}
\maketitle

\begin{abstract}
In the modern world, the amount of visual data recorded has been rapidly increasing. In many cases, data is stored in geographically distinct locations and thus requires a large amount of time and space to consolidate. Sometimes, there are also regulations for privacy protection which prevent data consolidation. In this work, we present federated implementations for object detection and recognition using a federated Faster R-CNN (FRCNN) and image segmentation using a federated Fully Convolutional Network (FCN). Our FRCNN was trained on 5000 examples of the COCO2017 dataset while our FCN was trained on the entire train set of the CamVid dataset. The proposed federated models address the challenges posed by the increasing volume and decentralized nature of visual data, offering efficient solutions in compliance with privacy regulations.
\end{abstract}


\section{Introduction}
\label{sec:intro}

Recently, the increase in applications driven by data has been accompanied by a growing concern for safeguarding data privacy. This has resulted in a growing interest in privacy-preserving machine learning algorithms. In federated learning, disparate parties collectively train a machine learning model without sharing training data \citep{FederatedLearningWithNon-IIDData}. This approach is especially useful in scenarios that involve sensitive or geographically dispersed data. Raw data remains secure, staying local to the individual federates while still contributing to the global training process \citep{zhang2023survey}. 

Many fields have restrictions in place limiting the types of data that can be shared. This is an especially important factor to consider when applied to defense purposes, as there may be inherent differences in classification levels of information being used on each local server that must be accounted for during the training process \citep{Demertzis,10.1117/12.2526661}. By utilizing the concepts set forth by FML, the secret levels of each federate can be maintained during training without the fear of Top Secret, Secret, Confidential, CUI or SCI from being spilled. 

Additionally, constraints on aggregating healthcare records exist due to H.I.P.A.A (or G.D.P.R in the EU), as patient data must be maintained securely \citep{Topaloglu_Morrell_Rajendran_Topaloglu_2021,elkordy2022privacy,annapareddy2023fairness,ramakrishnan2020compliant}. This necessitates the implementation of security standards for transferring information among different federates to comply with the standards set forth by the medical field \citep{info:doi/10.2196/41588,10.1145/3412357}. FML is a viable solution to this problem as each patient record stays local to its respective federate.

Smartphone manufacturers such as Apple and Google continually refine their machine learning models for object detection, recognition, and segmentation, driven by the abundance of user-generated data at their disposal. This research demonstrates the potential of federated learning in this context, enabling these companies to optimize their training processes locally while actively participating in global model advancements \citep{Ek_2022,hard2019federated,federated-personalization}. By adopting federated learning principles, smartphone manufacturers can increase both the efficiency and security of their model development, ensuring the confidentiality and integrity of user data.

Horizontal federated learning can also improve the accuracy of the trained model.  Since consolidating our data is no longer an issue, we are able to easily add more sources of data into our training cycle.  This practice tends to increase model accuracy since the training data is likely to be more robust.  Since the federates can be located in different locations geographically, data that was sourced from a range of environments is easier to include. Including data from geographically dispersed origins also comes with the benefit of having greater odds of encapsulating multiple settings, increasing the robustness of the model.

In this paper, we present our implementations of object detection, recognition, and segmentation using the Federated AI Technology Enabler (FATE) framework \citep{FATE}. Our federated object detection and recognition uses an FRCNN while our image segmentation model uses an FCN.

\section{Background}
\label{sec:Background}

\subsection{Federated Learning}

Federated machine learning (FML) is described by \cite{yang2019federated} as the definition of N agents that maintain some form of data \((A_{1}, A_{2}, …, A_{n})\) that all hold the same desire to develop a machine learning model that has access to each of their respective data pools \((D_{1}, D_{2}, …, D_{n})\). These data pools and the means by which they can be operated on can be separated into two categories, \textbf{Horizontal} and \textbf{Vertical} \citep{pmlr-v162-makhija22a}, discussed in sections \ref{sec:horizontal} and \ref{sec:vertical} respectively. One of the earliest federated learning implementations was able to jointly train ML models on Android devices by using a secure aggregation schema that protected user privacy \citep{yang2019federated}. Each of these different users, their data, as well as the server they operate from can be denoted as a federate \((F_{1}, F_{2}, …, F_{n})\), that of which serves all pertinent information necessary for training the model \(M_{FED}\).  

There are special security protocols associated with developing a federated model that is based on the separate training of models local to each federate \((M_{1}, M_{2}, …, M_{n})\) \citep{yang2019federated}. During this process, \(M_{1}, M_{2}, …, M_{n}\) have access to their own training weights and data references, but not to those of any other federate. In the same token, \(M_{FED}\) does not directly have access to the data associated with each federate, but does have a means of acquiring pertinent model information such as weights and layers respective to each federate. This allows every agent as well as the central server training \(M_{FED}\) to support their own private intelligence without any leaks occurring during the process. 

\begin{algorithm}
\caption{Client Training Process}
\label{alg:client}
\begin{algorithmic}[1]
\Procedure{ClientTrain}{$\Theta, D$}
    \For{$e \gets 1$ to $E$}
        \For{each $b$ in $D$}
            \State $\nabla \gets \text{compute\_gradients}(\Theta, b)$
            \State $\Theta \gets \Theta - \alpha \cdot \nabla$
        \EndFor
    \EndFor
    \State \textbf{return} $\Theta$
\EndProcedure
\end{algorithmic}
\end{algorithm}

\begin{algorithm}
\caption{Federated Averaging at Server}
\label{alg:fedavg}
\begin{algorithmic}[1]
\Procedure{FedAvg}{$\Theta, \{\Theta_i\}$}
    \State $w \gets 0$
    \State $n \gets \text{len}(\{\Theta_i\})$ \Comment{Number of local models}

    \State $\bar{\Theta} \gets \frac{1}{n} \sum_{i=1}^{n} \Theta_i$

    \State $\Theta \gets \bar{\Theta}$
\EndProcedure
\end{algorithmic}
\end{algorithm}

\begin{figure}
    \centering
    \includegraphics[scale=2,width=0.8\textwidth]{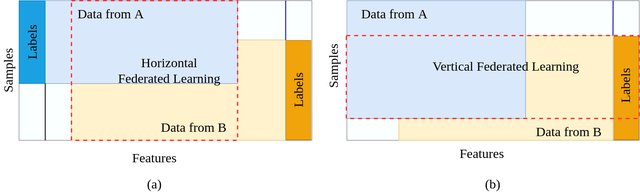}
    \caption{Data partitioning of horizontal and vertical federated learning}
    \label{fig:fedtypes}
\end{figure}

\subsection{Horizontal Federated Learning}
\label{sec:horizontal}
When considering model optimization and data privacy, there are a few approaches that can be integrated into a federated learning environment. Horizontal Federated Machine Learning is a viable approach when data is distributed across multiple federates, with each federate encompassing all relevant features and training labels associated with a specific data sample \citep{yang2019federated}, shown in figure \ref{fig:fedtypes}.a. This approach has seen the most significant focus regarding its development, causing its application to represent the most matured theories pertaining to FML \citep{9874186}. The system is oriented around a shared feature space, that of which references the encrypted data found on the individual federates in order to establish parameters and weights respective to the global model. The most common approach is the Federated averaging (FedAvg) algorithm, a decentralized machine learning approach where model parameters are trained across multiple local devices, and periodically aggregated to construct a global model, enabling collaborative learning without sharing raw data. A typical FedAvg implementation is shown in algorithm \ref{alg:fedavg} while a typical client's training process is shown in algorithm \ref{alg:client}.

\subsection{Vertical Federated Learning}
\label{sec:vertical}
Although not as common as horizontal federated learning due to its increased complexity, vertical federated learning encompasses a unique set of problems and use cases that must be mentioned. Vertical tasks operate on datasets whose samples share some common items in their feature spaces, as well as some within their data space \citep{10.1145/3514221.3526127,liu2022vertical}. A visualization of this distribution is shown in figure \ref{fig:fedtypes}.b.
Such a problem poises a set of challenges that do not arise when working with horizontal learning, as the data is not necessarily aligned. As such, vertical operations seek to uphold the same privacy applications all the while developing gradients and weights based upon potentially unrelated data points.

\subsection{Federated Computer Vision}
The main focus of this paper is the adaptation of computer vision in FATE. Due to its ability to effectively transfer data between multiple federates all the while maintaining secrecy amongst clients, it makes for an excellent environment that is more than capable of operating on image based problems. There exist some libraries and functionalities for the respective processes in the form of FedVision \citep{liu2020fedvision}. It is necessary, however, to expand on the existing use cases. In its current state, the package supports YOLOV3 object detection as well as some MNIST examples. FedVision is not interoperable with FATE, which is unfortunate as FATE is a much more robust platform.

\begin{figure}
    \centering
    \includegraphics[scale=2,width=0.45\textwidth]{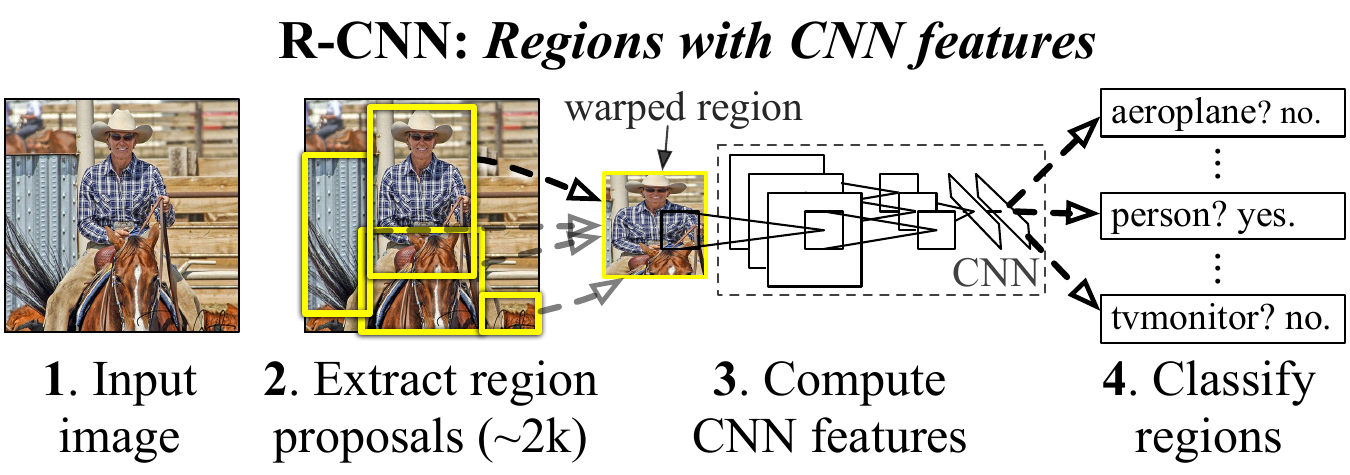}
    \caption{Architecture of an RCNN, as proposed in \cite{7112511}}
    \label{fig:rcnn}
\end{figure}

\subsubsection{Object Detection and Region-Based Convolutional Neural Networks}
Object Detection and Recognition is one of the most common computer vision problems \citep{zhao2019object}.  This process involves the acknowledgement of particular classes within a dataset, such as an automobile or pedestrian, as well as assigning them a numeric label. Subsequently, a series of bounding boxes are generated by the machine in order to establish a means of visualizing the data. These operations work in tandem to create a region based solution that provides an understanding of object location and the respective certainty tied to it.

Region Based Convolutional Neural Networks (RCNN), introduced in \cite{7112511}, are some of the most common models used for object detection and recognition \citep{SONG2023150,10198844}.
An RCNN creates a series of bounding boxes based on a feature pyramid network. These features are then swept by a region proposal network by a series of convolutions, those of which build a series of parameters and weights that are propagated throughout the remainder of the network. The features in this network are built in a pyramid to inverse pyramid fashion, prompting the data being worked on to rapidly decrease in feature size and subsequently being rebuilt to the original size. Doing so prompts minimal data loss, all the while maintaining ideal performance times. At the end of this process, the resulting data consists of the original frame or image overlaid by a series of best fit bounding boxes that annotate the region in which an object is located, as well as the class it is associated with. An overview of the RCNN is illustrated in figure \ref{fig:rcnn}

\begin{figure}
    \centering
    \includegraphics[scale=2,width=0.45\textwidth]{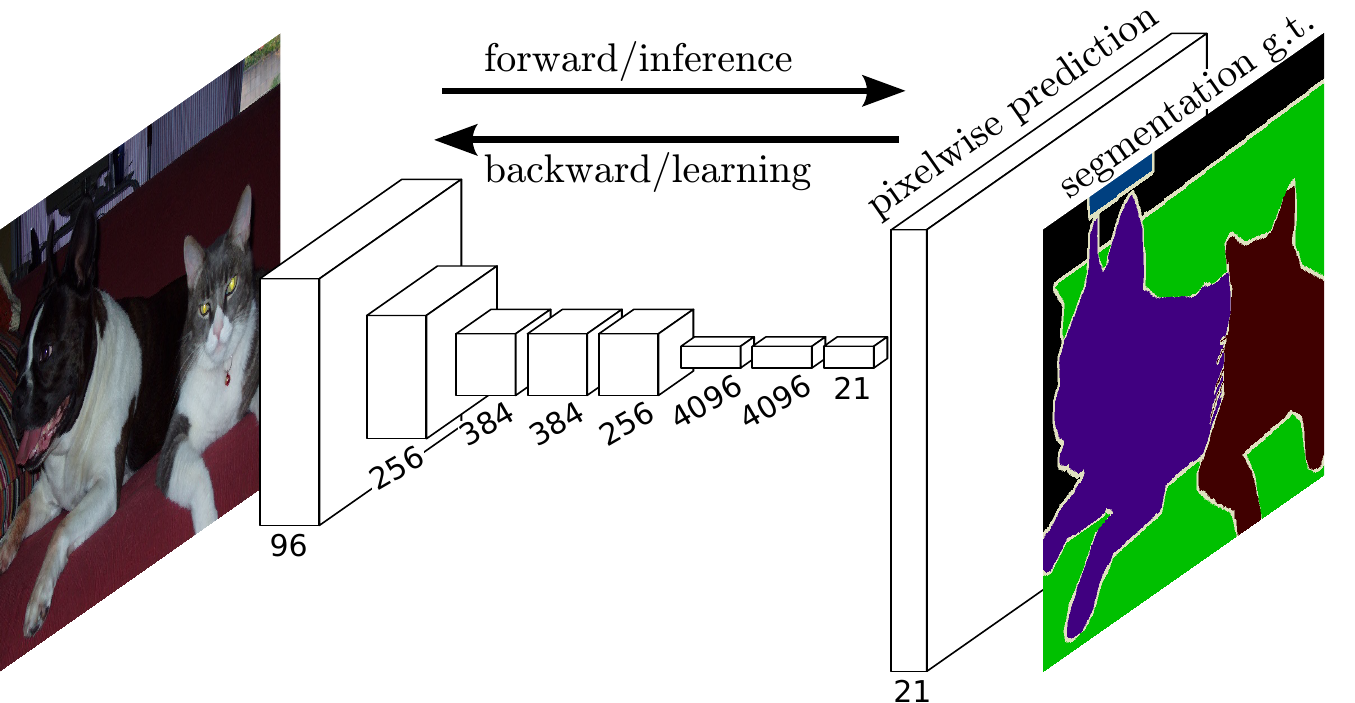}
    \caption{Architecture of an FCN, as proposed in \cite{long2015fully}}
    \label{fig:fcn}
\end{figure}

\subsubsection{Image Segmentation and Fully Convolutional Neural Networks}
\label{sec:segmentation}
Image segmentation is the process by which a sample image is broken down into a series of partitions that come in the form of regions \citep{minaee2020image}. Each of these sections can be assigned a class corresponding to an assessed ground truth, that of which acts as a label during deep learning training scenarios. The most common models trained on image segmentation problems are fully convolutional networks (FCN) which are used in in \cite{patravali20172d3d}, \cite{long2015fully}, and \cite{10227299}.

FCNs maintain the same hierarchal, end-to-end approach as standard CNN approaches all the while providing additional freedoms in regard to sample choices. They are inherently dynamic in their applications and can operate on samples of different pixel sizes. In order to do so, the model schema avoids the use of Dense layers as the standard for receiving inputs and outputs from previous convolutions, instead opting to use smaller convolutional layers to act as filters \citep{long2015fully}. By doing this, the natural limitations of singular input filters do not apply to the model, allowing for a much more diverse data pool that can change in size without any repercussions. This means that datasets containing samples that are 720 x 960 pixels can be used alongside those that host images that are 512 x 512 \citep{zheng2021rethinking}. The benefit to doing so is a direct increase in data availability, allowing models to obtain higher precision and accuracy, as well as uphold a larger number of use cases as a result. An additional benefit to utilizing interconnected convolutions is the inherently smaller parameter count as compared to intermittent dense layers. This schema avoids any data loss that may be attributed to mismanagement of weights between convolutions, as well as prompts a faster and more efficient training process.

A common network structure when utilizing an FCN for image segmentation resembles the structure of an autoencoder, with the difference being that the final result is a pixel mapping of the original image that is ultimately built upon fully connected layers for feature mapping. By utilizing a custom, or prebuilt network such as VGGNet \citep{wang2015places205vggnet} to apply a series of convolutions to a batch of samples, one can obtain useful feature extractions that may be applied to a convolutional transpose network. This approach yields an effective solution for working with troublesome sample sizes all the while maintaining the same performance of previously proposed methods. An architecture of an FCN is shown in figure \ref{fig:fcn}.

\section{Data Sets}
\label{sec:data}
\subsection{COCO 2017}
\label{sec:coco}
In this study, we have used an implementation of the Common Objects in Context (COCO2017) dataset \citep{lin2015microsoft}.  This dataset contains 328,000 annotated images with 91 unique object classes, 80 of which appear in the annotated data.  The annotations describe the location and dimensions of bounding boxes as well as the class of each object in the image.

\subsection{CamVid}
\label{sec:camvid}
\begin{figure}
    \centering
    \includegraphics[scale=2,width=0.45\textwidth]{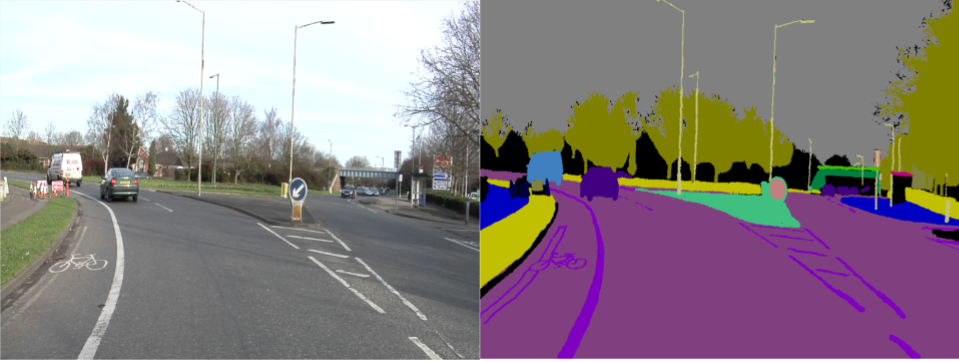}
    \caption{Left: Source Image | Right: Semantic Mask}
    \label{fig:seglabel}
\end{figure}

To determine the effectiveness of federated image segmentation, the Cambridge-Driving Labeled Video Database (CamVid), as proposed in \cite{BrostowFC:PRL2008} and \cite{BrostowSFC:ECCV08}, was inputted in to our federated FCN. This collection of information consists of 367 training, 101 validation, and 233 testing pairs of imagery, wherein each pair consists of both the original image as well as its corresponding ground truth. The labelled data comes in the form of a semantic mask that provides highlighted regions that overlay the source image, wherein these locations are attributed to a particular class. An example of an image-label pair is visualized in Figure \ref{fig:seglabel}. For visualization purposes, each category maintains a unique identifying color in order to differentiate between all of the possible items located in a single image.

\section{Implementation details}
\subsection{Federated Technology Enabler (FATE)}
An important consideration when developing a federated learning environment is the scalability and communication mandated by the federates. A common method utilized in the past comes in the form of distributed learning. This is an open-source approach that introduces a series of data silos that maintain unique encryption services that seek to preserve data privacy. However, this system is inherently limited to small scale environments, only being applicable in scenarios most often pertaining to research or personal use. Regardless of machine learning practices, this approach has significant security issues that do not align with data privacy compliance standards \citep{Liu_2022}. As a result, applying such systems in larger industries becomes much more difficult due to the inherent complications that come with them, causing their implementations to be quite limited.

The Federated Technology Enabler (FATE) is an open source framework for federated learning that resolves many of the issues enumerated above \citep{FATE,Li_2023}. This platform acts as an AI learning ecosystem capable of applying FML practices in a way that better accounts for network size and security concerns as compared to standard distributed learning methods.

The FATE architecture consists of a variety of machine learning applications that coincide with the needs for federation and security protocols set forth by the respective working environment. These algorithms work in conjunction with an EggRoll distribution framework to carry out advanced computations in the network. The system promotes key concepts that are integral to a healthy FML system, maintaining privacy amongst all federates involved in the machine learning process. All intermittent results such as gradients and weights \citep{254465}, as well as necessary computations, are executed under a set encryption service, wherein a single federate’s exposure to any other parties within the framework only occurs once the model is fully trained. At this stage, all federates are updated with the final result but still lack access to any particular information associated with the other parties. 

For horizontal federated learning, training is performed identically across the different federates. Thus, the performance is lossless and all models achieve roughly the same performance without exposure to sensitive data.

\subsection{Modifications to FATE}
\subsubsection{Data Loaders}
For our FRCNN and image segmentation training, we designed two federated dataloader objects. The first dataloader was structured to encapsulate COCO labels, ensuring that the initial element returned represented an image, while the subsequent element corresponded to the COCO label. To elaborate on the COCO format, each label includes information about the image, such as its dimensions, file name, and a list of annotations. Annotations consist of details about objects present in the image, including the category, segmentation mask, and bounding box coordinates.

The second dataloader for image segmentation was more straightforward. It provided an image along with a segmentation map of the label. In this case, the segmentation map utilized a class-to-RGB dictionary provided by the CamVid dataset to assign each pixel to its respective class. The original code in the FCN implementation that we used, as presented in \cite{FCNgithub}, did not use vectorization. It instead preprocessed the data by looping through each pixel in the image, then saved the converted data into a numpy array. As this is not useful in a federated environment, we decided to reimplement this code to provide a vectorized implementation that quickly maps the appropriate values.

\subsubsection{Federated Average Trainers}

 Several modifications were introduced in the implementation of federated averaging within the FATE framework. In both our segmentation and object detection trainer, we had to disable the fate logging metrics. This is because FATE is not configured to handle metrics aside from regression and classification.

 For our image segmentation model, we changed our loss to BCEWithLogitsLoss(). Although small syntax changes had to be made to work with our FCN, the core code for FedAvg remained mostly the same.
 
 In the case of the FRCNN model, the losses pertinent to training were obtained directly from the torchvision FRCNN model. The losses (box\_classifier, box\_fpn\_regulation, classifier, and objectness) were simply aggregated to form a unified loss during training. These losses are discussed in more detail in section \ref{sec:frcnnmethods}.

 \section{Methods}
\label{sec:methods}
\begin{figure}
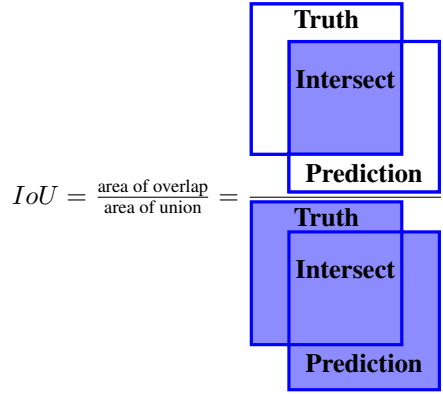

\centering
$IoU = \frac{\text{area of overlap}}{\text{area of union}}=
\frac{
    \tikz{
        \fill[draw=blue, very thick, fill=blue!45] (0,0) rectangle (2,2) (0.5,-0.5) rectangle (2.5,1.5);
        \fill[draw=blue, very thick, fill=white, even odd rule] (0,0) rectangle (2,2) (0.5,-0.5) rectangle (2.5,1.5);
        \node at (1, 1.8) {\textbf{Truth}};
        \node at (1.25, 1) {\textbf{Intersect}};
        \node at (1.5, -.25) {\textbf{Prediction}};
        }
    } 
    {
        \tikz{
            \fill[draw=blue, fill=blue!45, very thick] (0,-.1) rectangle (2,1.8) (0.5,-.7) rectangle (2.5,1.4);
            \node at (1, 1.6) {\textbf{Truth}};
            \node at (1.25, .9) {\textbf{Intersect}};
            \node at (1.5, -.35) {\textbf{Prediction}};
            }}$
\caption{IoU Calculation}
\label{fig:iou}
\end{figure}

Prior to model training, the data was split into two equally sized sets.  These different sets represent unique data silos that could work collaboratively in the training process.  We began by training on each set locally and tracking the loss and intersection over union (IoU) of the sets when compared to the ground truth.  IoU is a common metric used to track the accuracy of an object detection or segmentation model.  It is calculated by dividing the area of overlap between two bounding boxes by the area of their union. This results in a value between 0 and 1, where 1 represents a perfect overlap. We then trained on both sets in a federated manner and compared the results.  A visualization of the IoU metric is shown in figure \ref{fig:iou}.

\subsection{Federated FRCNN}
\label{sec:frcnnmethods}
After splitting the data into two sets, silo A and silo B, we passed each silo into a Faster Region-Based Convolutional Neural Network (FRCNN) introduced in \cite{ren2016faster}.  This consists of a feature pyramid network (FPN) which predicts points of interest, or features, in the image and draws a bounding box around them.  The image is then sent to a region proposal network (RPN) which evaluates the loss and modifies the weights of the network.

Once the models were done training on the silos locally we were able to visualize the loss and IOU over time.  These metrics give us an idea of how well our model is doing.  When doing object detection IOU is commonly used to track the accuracy since it quantifies how similar the predicted result is compared to the ground-truth by representing the ratio of overlapping area in the boxes compared to the entire area.

Our implementation of FRCNN returns four loss values during the training process: $loss\_objectness$, $loss\_bbox\_reg$, $loss\_bbox$, and $loss\_classifier$, each corresponding to a different aspect of the prediction as seen in \cite{ren2016faster}. $Loss\_objectness$ represents the objectness score of the predictions which is a way of quantifying the likelihood that the object belongs to a specific object class vs the background.  $Loss\_bbox\_reg$ quantifies how similar a given predicted bounding box is to the closest ground truth. 
Both of these values are calculated within the RPN.  $Loss\_bbox$ is used to measure how tight the predicted bounding boxes are to the ground truth object.  Lastly, $loss\_classifier$ is used to measure the correctness of each predicted label.

Once we obtained our metrics for training locally we then compared the metrics to a federated environment.  To achieve this we put the data from silo A on one federate and silo B on another with the goal of increasing IOU due to the collaborative learning process.

\subsection{Federated Image Segmentation}
\label{sec:segmethods}
To compare and contrast the results of standard modelling algorithms and those that are federated, the testing process was separated into two groups: one federated and another non-federated. The non-federated procedure takes into account the data splitting capabilities of the other test, thus creating two local models that were trained on partitions of equal size from the CamVid dataset. Here, the models were to be trained on roughly 183 image and label pairs in order to simulate a "federated" environment. The second test involves the true federated model, which consists of two federates that maintain access to the same split of data seen in the first experiment. Simulating the systems in this way provides insight to the benefits and limitations of the federated approach amid the additional networking and security capabilities seen by the architecture. In both experiments, metrics for pixel accuracy, IoU, and loss were obtained in order to determine performance.

\section{Results}
\label{sec:results}
\subsection{Federated FRCNN}
\label{sec:frcnnresults}

\begin{figure}
    \centering
    \includegraphics[scale=2,width=0.45\textwidth]{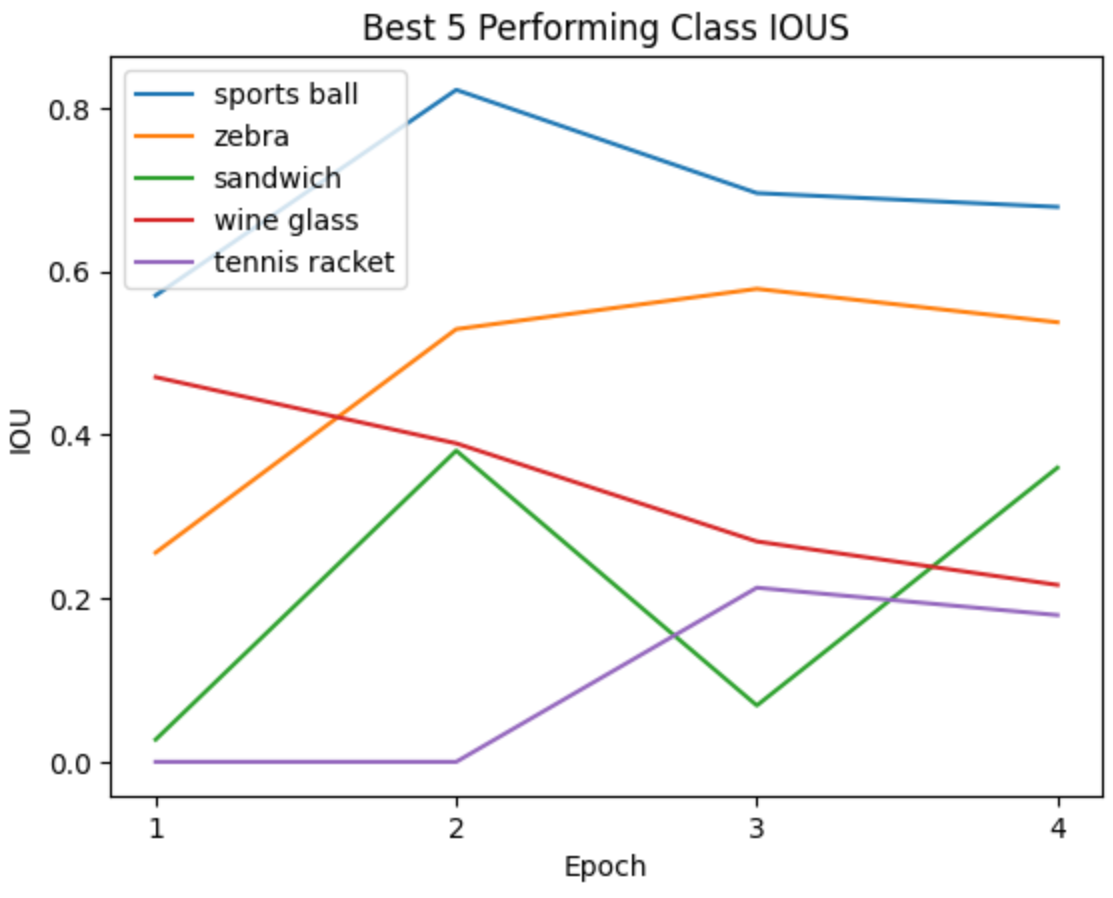}
    \caption{Top 5 IOU of FRCNN}
    \label{fig:top5}
\end{figure}

In this section, we present the results during the training of our object detection model. Figure \ref{fig:top5} shows the performance of the top highest performing 5 classes over 4 epochs.

Our results demonstrate the viability and potential benefits of implementing federated object detection and recognition in a decentralized environment. The IOU of the top 5 classes ranged from 20\% to 70\%. The overall IOU across all classes was around 10\% which is likely due to class imbalance of the COCO dataset and the lack of training examples for each class as we trained on 5000 images and had 80 classes.

\subsection{Federated Image Segmentation}

\begin{figure}
    \centering
    \begin{subfigure}{.5\textwidth}
        \centering
        \includegraphics[width=0.8\linewidth]{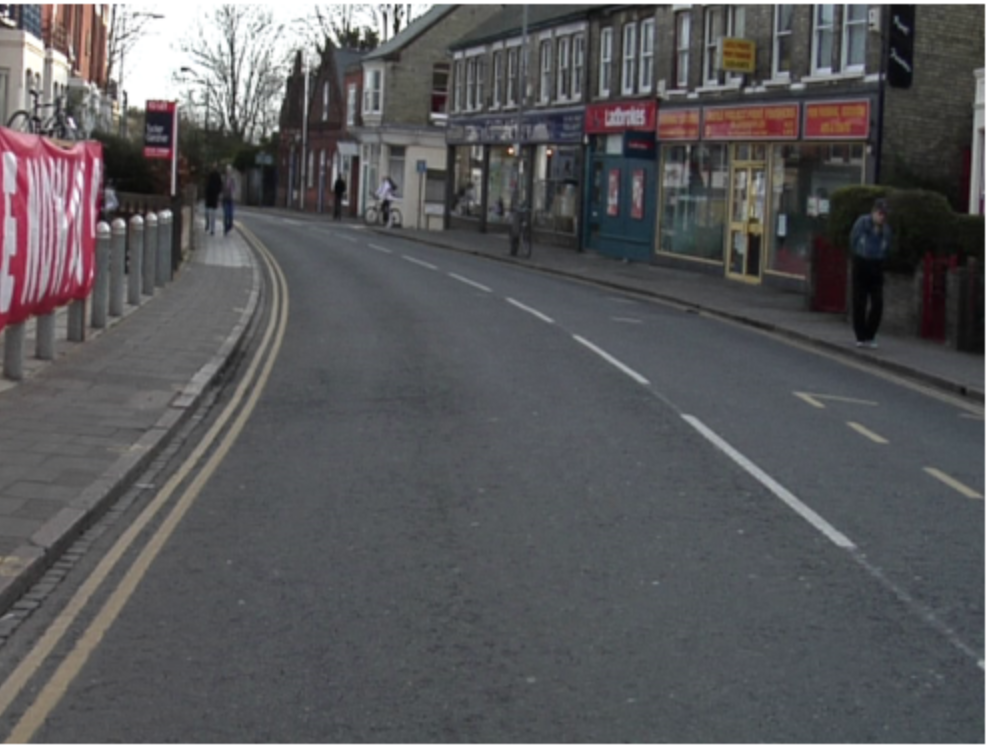}
        \caption{Source}
        \label{fig:seg_res1}
    \end{subfigure}%
    \begin{subfigure}{.5\textwidth}
        \centering
        \includegraphics[width=0.8\linewidth]{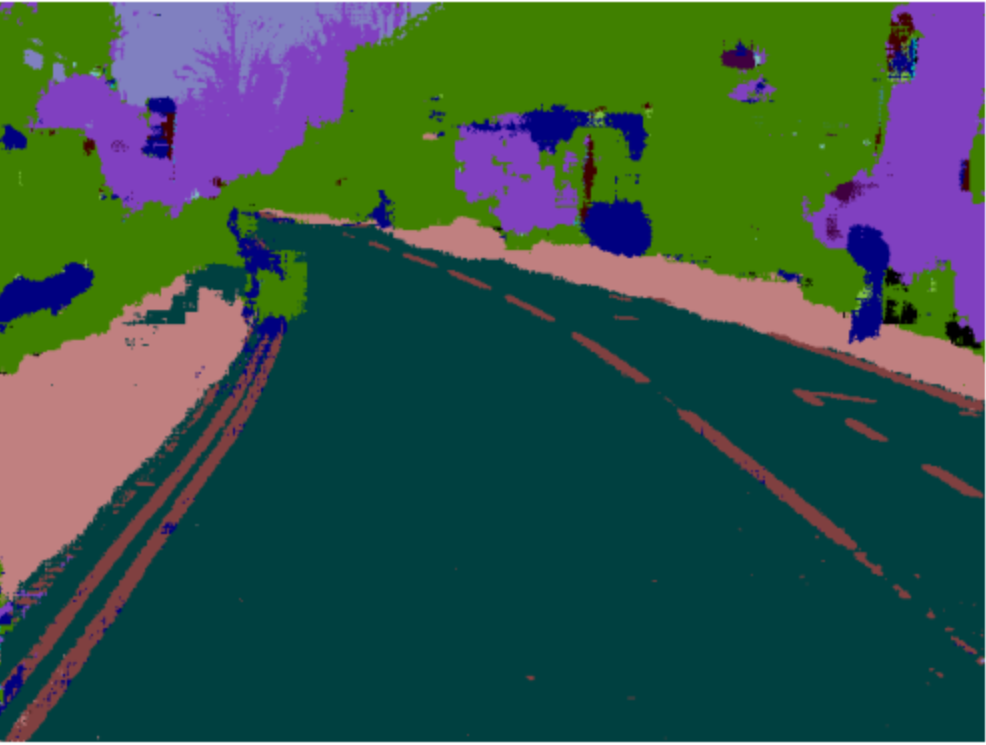}
        \caption{Segmentation}
        \label{fig:seg_res0}
    \end{subfigure}
    \caption{Result of a testing sample from our federated FCN}
    \label{fig:seg_res}
\end{figure}

\begin{figure}
    \centering
    \begin{subfigure}[t]{0.5\textwidth}
        \centering
        \includegraphics[scale=2,width=0.9\textwidth]{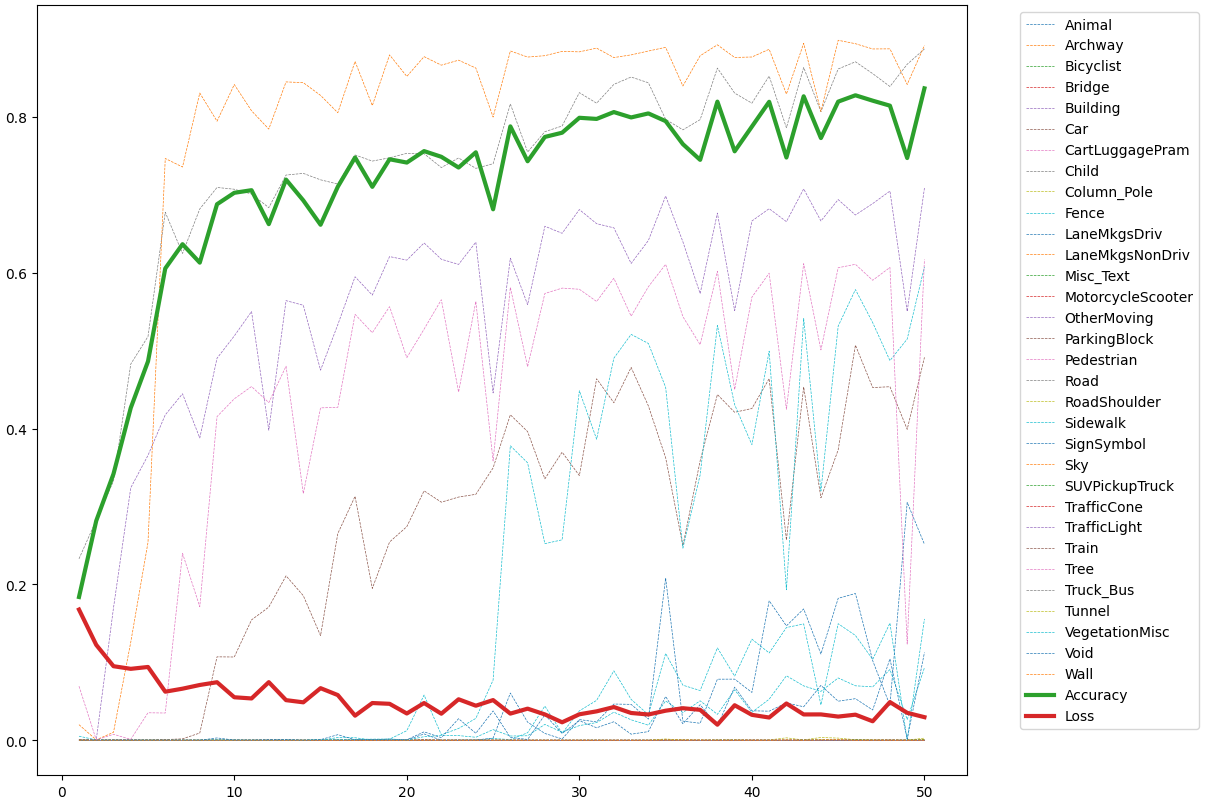}
        \caption{Performance metrics of a local model trained on federate A's data.}
    \end{subfigure}%
    ~ 
    \begin{subfigure}[t]{0.5\textwidth}
        \centering
        \includegraphics[scale=2,width=.9\textwidth]{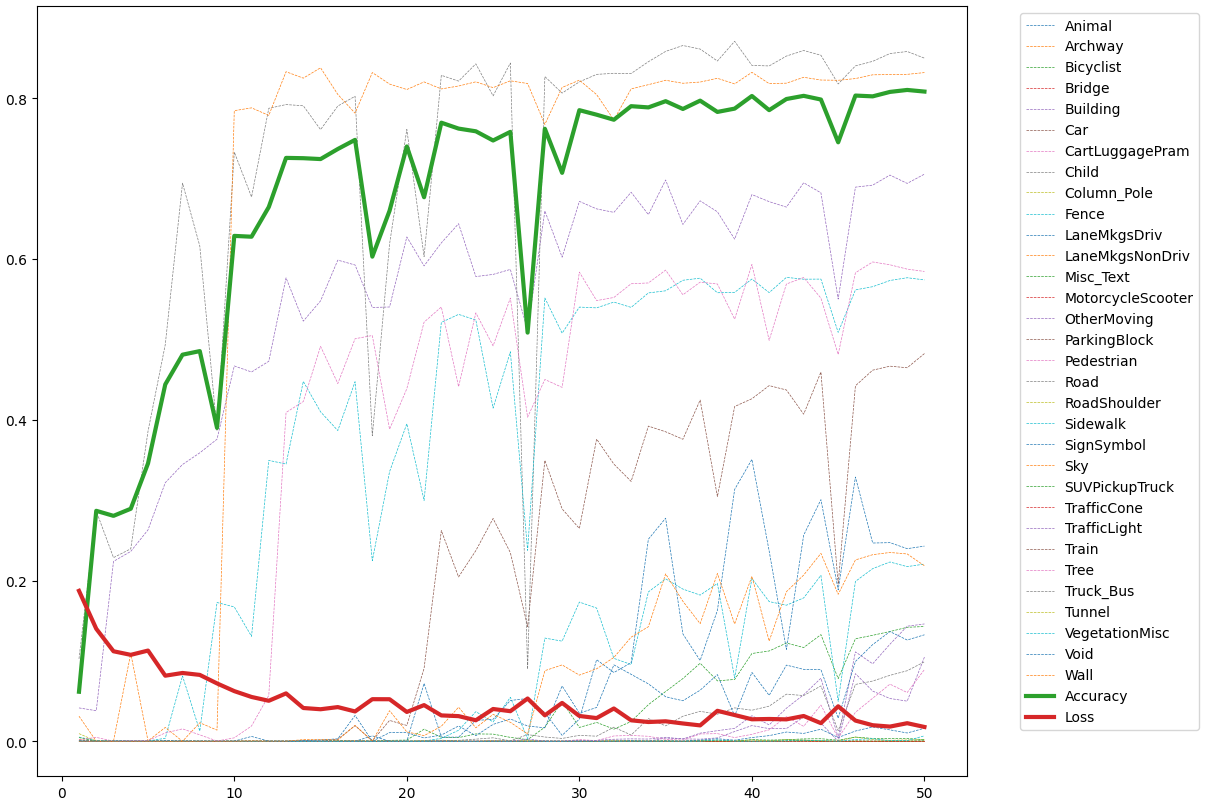}
        \caption{Performance metrics of a local model trained on federate B's data.}
    \end{subfigure}
    \centering
    \begin{subfigure}[t]{1\textwidth}
        \centering
        \includegraphics[scale=2,width=0.45\textwidth]{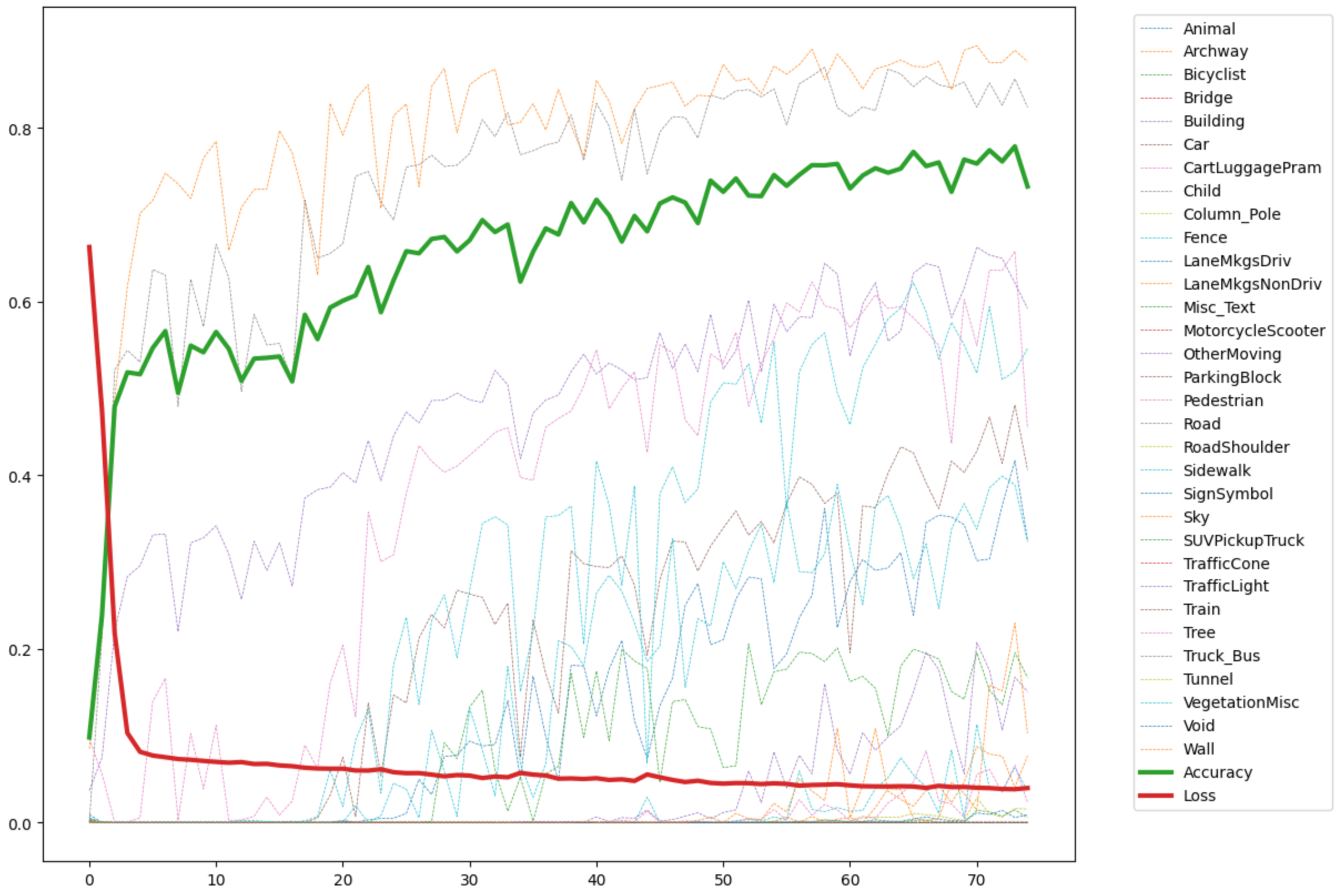}
        \caption{Performance metrics of the federated model.}
    \end{subfigure}
    \caption{IoU: dashed, Loss: red, Pixel Accuracy: green.}
    \label{fig:segres}
\end{figure}

The local models in this experiment reached a solution and converged after training for 50 epochs, whereas the federated model required 75 in order to reach proper conclusions. The performance of each can be seen in figure \ref{fig:segres}, that of which provides detailed graphical analyses of their capabilities throughout their training iterations. Being that IoU was class specific, the metric for each label had to be plotted individually. This is done in order to combat the presence of niche classes that the dataset does not effectively provide, and make it apparent that more consistent items such as roads and buildings see significant improvement. 

From these plots, it can be seen that the local models perform relatively the same with respect to their exposed data silos, converging to a pixel accuracy of 80\%. In reference to the IoUs determined by the system, it can be seen that notable classes such as buildings and roads approach values ranging from 75\% to 85\%, wherein more niche items such as car doors range anywhere from 5\% to 50\% depending on the provided samples. This explains the minute differences seen between the two local models operating on split data, wherein each is exposed to some items more frequently than others and are able to build the corresponding weights. 

In comparison, the federated model converged at an accuracy of 75\%, which is slightly lower than that of the first experiment. Although this slight decrease in performance brings into question the integrity of a truly federated approach, it is important to consider that the architecture being implemented is being operated in a much different scenario. In this experiment, the data hosted on each federate is shared with each other through a secure encryption schema that allows each to reference sensitive information without exposing it outside the network. This also prevents the need to directly share data with different locations amid the ability to directly reference the information stored on the parties within the system without ever having real access to it. A final consideration to make is the inherently small size of the CamVid dataset, that of which does not boast enough data in order to fully utilize the aforementioned benefits of a federated algorithm. Finally, the IoU scores achieved by the federated model are higher than those of the local models, having more prominent features maintaining 85\% to 95\%, and niche classes ranging from 5\% to 60\%. 

\section{Discussion \& Future work}
\label{sec:discussion}

The results showcased in the previous sections highlight the effectiveness and challenges of federated object detection and recognition, as well as federated image segmentation.

The FRCNN experiments demonstrated the benefits of collaborative learning across decentralized nodes. The federated approach consistently outperformed local trials after a specific number of epochs, showcasing the potential of leveraging diverse data sources for enhanced model generalization. The accelerated improvement in the object classifier within the federated environment emphasizes the synergistic effect of pooling knowledge from various contributors. The consistent loss convergence underlines the robustness of federated learning, suggesting that this approach is well-suited for tasks requiring large and diverse datasets.

Challenges such as communication overhead, model synchronization, and heterogeneity of data distributions need further exploration and optimization. Addressing these challenges will be crucial for ensuring the efficiency and scalability of federated object detection and recognition.

The federated image segmentation experiments revealed both strengths and challenges. The federated model required a longer training time compared to local models, raising considerations about the trade-offs between model accuracy and resource constraints. The performance analysis indicated that local models excelled in capturing features within their exposure, while the federated model demonstrated a better understanding of shared knowledge among nodes.

The variability in performance across different classes highlighted challenges associated with imbalanced datasets, particularly for niche classes with limited representation. The privacy-preserving nature of federated learning, enabled by encryption schemas, is a notable advantage, especially in scenarios where data privacy is paramount.

Federated Learning has evolved to support Large Language Models (LLMs), marking a significant advancement. Specifically, FATE-LLM uses low rank adaptations to train LLMs \citep{fan2023fatellm}. Many of these training algorithms can be used for Vision Transformers (ViTs). ViTs offer notable advantages, especially in tasks like object detection, recognition, and image segmentation. The self-supervised training inherent in ViTs can enhance the model's ability to learn hierarchical features from distributed datasets.

Additionally, the discussion should extend to exploring vertical implementations of these algorithms. Progress has already been made in image segmentation, indicating the feasibility and potential success in extending these approaches to other related tasks.

\bibliographystyle{apalike}
\bibliography{egbib}

\begin{thebibliography}{}

\bibitem[Annapareddy et~al., 2023]{annapareddy2023fairness}
Annapareddy, N., Preston, J., and Fox, J. (2023).
\newblock Fairness and privacy in federated learning and their implications in healthcare.

\bibitem[Brauneck et~al., 2023]{info:doi/10.2196/41588}
Brauneck, A., Schmalhorst, L., Kazemi~Majdabadi, M.~M., Bakhtiari, M., V{\"o}lker, U., Baumbach, J., Baumbach, L., and Buchholtz, G. (2023).
\newblock Federated machine learning, privacy-enhancing technologies, and data protection laws in medical research: Scoping review.

\bibitem[Brostow et~al., 2009]{BrostowFC:PRL2008}
Brostow, G.~J., Fauqueur, J., and Cipolla, R. (2009).
\newblock Semantic object classes in video: A high-definition ground truth database.
\newblock {\em Pattern Recognition Letters}, 30(2):88--97.

\bibitem[Brostow et~al., 2008]{BrostowSFC:ECCV08}
Brostow, G.~J., Shotton, J., Fauqueur, J., and Cipolla, R. (2008).
\newblock Segmentation and recognition using structure from motion point clouds.
\newblock In {\em ECCV (1)}, pages 44--57.

\bibitem[Cirincione and Verma, 2019]{10.1117/12.2526661}
Cirincione, G. and Verma, D. (2019).
\newblock {Federated machine learning for multi-domain operations at the tactical edge}.

\bibitem[Demertzis et~al., 2023]{Demertzis}
Demertzis, K., Kikiras, P., Skianis, C., Rantos, K., Iliadis, L., and Stamoulis, G. (2023).
\newblock Federated auto-meta-ensemble learning framework for ai-enabled military operations.

\bibitem[Ek et~al., 2022]{Ek_2022}
Ek, S., Portet, F., Lalanda, P., and Vega, G. (2022).
\newblock Evaluation and comparison of federated learning algorithms for human activity recognition on smartphones.
\newblock {\em Pervasive and Mobile Computing}, 87:101714.

\bibitem[Elkordy et~al., 2022]{elkordy2022privacy}
Elkordy, A.~R., Zhang, J., Ezzeldin, Y.~H., Psounis, K., and Avestimehr, S. (2022).
\newblock How much privacy does federated learning with secure aggregation guarantee?

\bibitem[Fan et~al., 2023]{fan2023fatellm}
Fan, T., Kang, Y., Ma, G., Chen, W., Wei, W., Fan, L., and Yang, Q. (2023).
\newblock Fate-llm: A industrial grade federated learning framework for large language models.

\bibitem[Fu et~al., 2022]{10.1145/3514221.3526127}
Fu, F., Xue, H., Cheng, Y., Tao, Y., and Cui, B. (2022).
\newblock Blindfl: Vertical federated machine learning without peeking into your data.

\bibitem[Gao et~al., 2023]{10227299}
Gao, F., Huang, H., Yue, Z., Li, D., Ge, S.~S., Lee, T.~H., and Zhou, H. (2023).
\newblock Cross-modality features fusion for synthetic aperture radar image segmentation.

\bibitem[Girshick et~al., 2016]{7112511}
Girshick, R., Donahue, J., Darrell, T., and Malik, J. (2016).
\newblock Region-based convolutional networks for accurate object detection and segmentation.

\bibitem[Hard et~al., 2019]{hard2019federated}
Hard, A., Rao, K., Mathews, R., Ramaswamy, S., Beaufays, F., Augenstein, S., Eichner, H., Kiddon, C., and Ramage, D. (2019).
\newblock Federated learning for mobile keyboard prediction.

\bibitem[Huang, 2018]{FCNgithub}
Huang, P.-C. (2018).
\newblock The easiest implementation of fully convolutional networks.
\newblock \url{https://github.com/pochih/FCN-pytorch}.
\newblock Accessed: December 16, 2023.

\bibitem[Li et~al., 2023]{Li_2023}
Li, Q., Wen, Z., Wu, Z., Hu, S., Wang, N., Li, Y., Liu, X., and He, B. (2023).
\newblock A survey on federated learning systems: Vision, hype and reality for data privacy and protection.

\bibitem[Lin et~al., 2015]{lin2015microsoft}
Lin, T.-Y., Maire, M., Belongie, S., Bourdev, L., Girshick, R., Hays, J., Perona, P., Ramanan, D., Zitnick, C.~L., and Dollár, P. (2015).
\newblock Microsoft coco: Common objects in context.

\bibitem[Liu et~al., 2022a]{9874186}
Liu, D., Bai, L., Yu, T., and Zhang, A. (2022a).
\newblock Towards method of horizontal federated learning: A survey.

\bibitem[Liu et~al., 2022b]{Liu_2022}
Liu, J., Huang, J., Zhou, Y., Li, X., Ji, S., Xiong, H., and Dou, D. (2022b).
\newblock From distributed machine learning to federated learning: a survey.
\newblock {\em Knowledge and Information Systems}, 64(4):885--917.

\bibitem[Liu et~al., 2021]{FATE}
Liu, Y., Fan, T., Chen, T., Xu, Q., and Yang, Q. (2021).
\newblock Fate: An industrial grade platform for collaborative learning with data protection.

\bibitem[Liu et~al., 2020]{liu2020fedvision}
Liu, Y., Huang, A., Luo, Y., Huang, H., Liu, Y., Chen, Y., Feng, L., Chen, T., Yu, H., and Yang, Q. (2020).
\newblock Fedvision: An online visual object detection platform powered by federated learning.

\bibitem[Liu et~al., 2022c]{liu2022vertical}
Liu, Y., Kang, Y., Zou, T., Pu, Y., He, Y., Ye, X., Ouyang, Y., Zhang, Y.-Q., and Yang, Q. (2022c).
\newblock Vertical federated learning.

\bibitem[Long et~al., 2015]{long2015fully}
Long, J., Shelhamer, E., and Darrell, T. (2015).
\newblock Fully convolutional networks for semantic segmentation.

\bibitem[Makhija et~al., 2022]{pmlr-v162-makhija22a}
Makhija, D., Han, X., Ho, N., and Ghosh, J. (2022).
\newblock Architecture agnostic federated learning for neural networks.

\bibitem[Minaee et~al., 2020]{minaee2020image}
Minaee, S., Boykov, Y., Porikli, F., Plaza, A., Kehtarnavaz, N., and Terzopoulos, D. (2020).
\newblock Image segmentation using deep learning: A survey.

\bibitem[Patravali et~al., 2017]{patravali20172d3d}
Patravali, J., Jain, S., and Chilamkurthy, S. (2017).
\newblock 2d-3d fully convolutional neural networks for cardiac mr segmentatin.

\bibitem[Paulik et~al., 2022]{federated-personalization}
Paulik, M., Seigel, M., Mason, H., Telaar, D., Kluivers, J., van Dalen, R., Lau, C.~W., Carlson, L., Granqvist, F., Vandevelde, C., Agarwal, S., Freudiger, J., Byde, A., Bhowmick, A., Kapoor, G., Beaumont, S., Áine Cahill, Hughes, D., Javidbakht, O., Dong, F., Rishi, R., and Hung, S. (2022).
\newblock Federated evaluation and tuning for on-device personalization: System design \& applications.

\bibitem[Pfitzner et~al., 2021]{10.1145/3412357}
Pfitzner, B., Steckhan, N., and Arnrich, B. (2021).
\newblock Federated learning in a medical context: A systematic literature review.

\bibitem[Ramakrishnan et~al., 2020]{ramakrishnan2020compliant}
Ramakrishnan, G., Nori, A., Murfet, H., and Cameron, P. (2020).
\newblock Towards compliant data management systems for healthcare ml.

\bibitem[Ren et~al., 2016]{ren2016faster}
Ren, S., He, K., Girshick, R., and Sun, J. (2016).
\newblock Faster r-cnn: Towards real-time object detection with region proposal networks.

\bibitem[Song et~al., 2023]{SONG2023150}
Song, P., Li, P., Dai, L., Wang, T., and Chen, Z. (2023).
\newblock Boosting r-cnn: Reweighting r-cnn samples by rpn’s error for underwater object detection.

\bibitem[Topaloglu et~al., 2021]{Topaloglu_Morrell_Rajendran_Topaloglu_2021}
Topaloglu, M.~Y., Morrell, E.~M., Rajendran, S., and Topaloglu, U. (2021).
\newblock In the pursuit of privacy: The promises and predicaments of federated learning in healthcare.

\bibitem[Wang et~al., 2015]{wang2015places205vggnet}
Wang, L., Guo, S., Huang, W., and Qiao, Y. (2015).
\newblock Places205-vggnet models for scene recognition.

\bibitem[Wang et~al., 2023]{10198844}
Wang, R., Luo, M., Feng, Q., Peng, C., and He, D. (2023).
\newblock Multi-party privacy-preserving faster r-cnn framework for object detection.
\newblock {\em IEEE Transactions on Emerging Topics in Computational Intelligence}, pages 1--12.

\bibitem[Yang et~al., 2019]{yang2019federated}
Yang, Q., Liu, Y., Chen, T., and Tong, Y. (2019).
\newblock Federated machine learning: Concept and applications.

\bibitem[Zhang et~al., 2020]{254465}
Zhang, C., Li, S., Xia, J., Wang, W., Yan, F., and Liu, Y. (2020).
\newblock {BatchCrypt}: Efficient homomorphic encryption for {Cross-Silo} federated learning.

\bibitem[Zhang et~al., 2023]{zhang2023survey}
Zhang, Y., Zeng, D., Luo, J., Xu, Z., and King, I. (2023).
\newblock A survey of trustworthy federated learning with perspectives on security, robustness, and privacy.

\bibitem[Zhao et~al., 2018]{FederatedLearningWithNon-IIDData}
Zhao, Y., Li, M., Lai, L., Suda, N., Civin, D., and Chandra, V. (2018).
\newblock Federated learning with non-iid data.

\bibitem[Zhao et~al., 2019]{zhao2019object}
Zhao, Z.-Q., Zheng, P., tao Xu, S., and Wu, X. (2019).
\newblock Object detection with deep learning: A review.

\bibitem[Zheng et~al., 2021]{zheng2021rethinking}
Zheng, S., Lu, J., Zhao, H., Zhu, X., Luo, Z., Wang, Y., Fu, Y., Feng, J., Xiang, T., Torr, P. H.~S., and Zhang, L. (2021).
\newblock Rethinking semantic segmentation from a sequence-to-sequence perspective with transformers.

\end{thebibliography}

\end{document}